%% file: egpaper_final.tex
\documentclass[10pt,twocolumn,letterpaper]{article}

\usepackage{iccv}
\usepackage{times}
\usepackage{epsfig}
\usepackage{graphicx}
\usepackage{amsmath}
\usepackage{amssymb}

\usepackage[breaklinks=true,bookmarks=false]{hyperref}

\iccvfinalcopy 

\def\iccvPaperID{****} 

\ificcvfinal\fi

\begin{document}

\title{Cascade Weight Shedding in Deep Neural Networks:\\
Benefits and Pitfalls for Network Pruning}

\author{Kambiz Azarian, Fatih Porikli\\
Qualcomm Technologies Inc.\\
San Diego, CA 92121, USA\\
{\tt\small \{kambiza, fporikli\}@qti.qualcomm.com}
}

\maketitle
\ificcvfinal\thispagestyle{empty}\fi

\begin{abstract}
    \input{LaTeX/sections/abstract.}
\end{abstract}

\section{Introduction}\label{sec:introduction}
    \input{LaTeX/sections/introduction}

\section{Related Work}\label{sec:related_work}
    \input{sections/related_work.tex}

\section{Methodology}\label{sec:method}
\input{sections/method.tex}

\section{Experiments}\label{sec:experiments}
\input{sections/experiments.tex}

\section{Future Directions}\label{sec:future_directions}
\input{LaTeX/sections/future_directions}

\section{Conclusion}\label{sec:conclusion}
\input{sections/conclusion.tex}

{\small
\bibliographystyle{ieee_fullname}
\bibliography{egbib}
}

\end{document}

%% file: LaTeX/sections/abstract..tex
We report, for the first time, on the cascade weight shedding phenomenon in deep neural networks where in response to pruning a small percentage of a network's weights, a large percentage of the remaining is shed over a few epochs during the ensuing fine-tuning phase. We show that cascade weight shedding, when present, can significantly improve the performance of an otherwise sub-optimal scheme such as random pruning. This explains why some pruning methods may perform well under certain circumstances, but poorly under others, e.g., ResNet50 vs. MobileNetV3. We provide insight into why the global magnitude-based pruning, i.e., GMP, despite its simplicity, provides a competitive performance for a wide range of scenarios. We also demonstrate cascade weight shedding's potential for improving GMP's accuracy, and reduce its computational complexity. In doing so, we highlight the importance of pruning and learning-rate schedules. We shed light on weight and learning-rate rewinding methods of re-training, showing their possible connections to the cascade weight shedding and reason for their advantage over fine-tuning. We also investigate cascade weight shedding's effect on the set of kept weights, and its implications for semi-structured pruning. Finally, we give directions for future research.

%% file: LaTeX/sections/introduction.tex
The ever-increasing demand for efficient deployment of deep neural networks has resulted in a myriad of pruning methods, whether unstructured \cite{han2015pruning, zhu2017prune, azarian2020learned}, structured \cite{he2017, vibnet} or semi-structured \cite{block-sparsity-mao, balanced-sparsity-yao}. Yet the majority of these methods ascribe, if only implicitly, a certain static characteristic to the pruning process, e.g., they require a post-pruning \emph{fine-tuning} phase with a small learning-rate for the weights to \emph{adjust}, or consider weight distributions to stay stationary throughout the pruning process. In this paper we show that at least for unstructured pruning and under certain conditions, this may not be true. More specifically we report, for the first time, on the cascade weight shedding phenomenon, where in response to pruning a small percentage of a network's weights, a large portion of the remaining is shed over a few epochs during the subsequent re-training. We also give some of the necessary conditions for cascade weight shedding's occurrence and networks prone to it.

Understanding the implications of the cascade weight shedding is very important. Not only networks prone to cascade weight shedding such as ResNet \cite{resnet}, Vgg \cite{vgg} and InceptionV3 \cite{inceptionv3} are in widespread use as either standalone networks or backbones for other networks (e.g., DeepLabV3 \cite{chen2018deeplab}, SSD \cite{liu2016ssd}), but also they are the networks of choice when investigating new pruning methods. The fact that other popular networks such as EfficientNet \cite{efficientnet} and MobileNetV3 \cite{mobilenetv3} are not prone, at least to the same degree, to cascade weight shedding makes this understanding even more important. Studying the weight shedding process also provides valuable insight into internal workings of unstructured pruning methods such as the iterative global magnitude pruning, i.e., GMP \cite{han2015pruning}. For example, while it is perceivable that the bottom, e.g., $10\%$ of a network's weights are spurious and redundant, it is not so clear why removing the bottom, e.g., $85\%$ (including the weights well above the median of the distribution) should still give a good performance. Studying cascade weight shedding, in addition, provides methods to speed-up the pruning process without sacrificing performance. Overall, studying the cascade weight shedding, by highlighting the high degree of dynamism involved in the pruning process, puts more emphasis on pruning and learning-rate schedules, and away from heuristics for identifying redundant weights.

Our contributions in this paper are the following:
\begin{itemize}
    \item We describe, for the first time, the cascade weight shedding phenomenon in deep neural networks. We also give some necessary conditions for its occurrence, and some of the networks prone to it, c.f., section \ref{subsec:gmp_shedding}.
    
    \item We show that cascade weight shedding, when present, can significantly improve the performance of an otherwise sub-optimal scheme, explaining why some methods perform well for some scenarios, but poorly otherwise, c.f., section \ref{subsec:rand_shedding}.
    
    \item We provide insight into the iterative GMP's internal workings, and demonstrate cascade weight shedding's potential for improving its accuracy, and reducing its computational complexity. In doing so, we highlight the importance of pruning and learning-rate schedules, c.f., sections \ref{subsec:rand_shedding} and \ref{subsec:exp_pruning}.
    
    \item We shed light on weight and learning-rate rewinding methods of re-training \cite{renda2020comparing}. In particular, we discuss their possible connections to cascade weight shedding, c.f., section \ref{subsec:weight_rewinding}. 
 
    \item We investigate cascade weight shedding's effects on the final set of kept weights, and its implications for semi-structured pruning, c.f., sections \ref{subsec:which_weights} and \ref{subsec:semi_struct}.

    \item Finally, we give a few directions for future research in section \ref{sec:future_directions}.
\end{itemize}

%% file: sections/related_work.tex
Several methods have been proposed for structured, unstructured and semi-structured pruning of deep neural networks, e.g., structured methods like \cite{he2017, lipruning} used layer-wise statistics to remove redundant channels from pre-trained layers. Other structured methods like Bayesian compression \cite{bayesiancompression}, VIBNets \cite{vibnet} and L1/L0-regularization \cite{srinivas2017sparse, christosl0} used gates to prune channels while training. Structured pruning methods, while accelerating inference and training of deep neural networks on general-purpose hardware, degrade the accuracy more significantly compared to unstructured pruning methods that remove individual weights. This latter type of pruning has been in use since $1989$ in the optimal brain damage \cite{obd} and optimal brain surgeon \cite{obs} papers, which removed individual weights based on Hessian information.

Global magnitude pruning, the most influential unstructured method of pruning, was introduced in \cite{han2015pruning} and later used in \cite{han2015deep} as part of a full model compression pipeline, where smaller weights were removed and the rest were fine-tuned afterwards. Magnitude-based methods, despite their simplicity, have been used in numerous subsequent works with good results, e.g., \cite{zhu2017prune} used a cubical polynomial pruning schedule and an exponentially decaying learning-rate schedule for gradual pruning of networks. Another application of GMP was by \cite{renda2020comparing} where weight and learning-rate rewinding schemes were used to achieve competitive pruning performances. More recently, \cite{azarian2020learned}, \cite{kusupati2020soft} and \cite{automatedpruningmanessi} used magnitude-based pruning together with different soft-threshold operators to learn per-layer thresholds through back-propagation. GMP has also been used for finding sub-networks, i.e., winning lottery tickets, that work as well as their original network in \cite{lotteryticket, deconstructinglottery}. 

Magnitude-based techniques, however, are not the only unstructured pruning methods, e.g., \cite{admm} and \cite{progressiveadmm} apply the alternating method of Lagrange multipliers, i.e., ADMM, to slowly coax a network into pruning weights, whereas \cite{molchanov2017, ullrichsws} utilize a variational Bayesian framework. Unstructured pruning methods, despite showing good accuracies at high pruning ratios, are not amenable to implementation on CPU and GPU and require special-purpose hardware for acceleration to be gained. This has motivated semi-structured methods of pruning where the sparsity is regularized to make it utilizable by general-purpose hardware. For example \cite{block-sparsity-mao, block-sparsity-narang} generalize GMP to either keep or prune entire blocks of adjacent weights, while \cite{balanced-sparsity-yao, balanced-sparsity-cao} induce balanced sparsity by independently pruning each block to the desired level using unstructured methods.     

We observe that while most methods operating on a pre-trained model required a post-pruning re-training phase, the majority assumed this to be a \emph{fine-tuning} operation with a very small learning-rate. Similarly, various gating methods assumed the network statistics, such as weight distributions, to remain stationary throughout the pruning process. These assumptions were also reflected in likening the pruning process to that of pruning unwanted branches from a tree, while keeping the rest \emph{intact}. In this work, we show that the pruning process is more dynamic than previously thought, i.e., that pruning a set of weights causes the network to shed other weights, which under certain scenarios may trigger a cascade event. Interestingly, while this phenomenon has not been reported in the literature before, its effects are discernible in the hindsight, e.g., \cite{han2015pruning} reported iterative GMP to outperform the single-shot variant, \cite{zhu2017prune} reported occasional near-catastrophic performance hits while pruning, \cite{renda2020comparing} reported weight and learning-rate rewinding methods to outperform fine-tuning and \cite{azarian2020learned, kusupati2020soft} showed that the keep-ratio time-traces of their schemes resembled an exponential-decay, despite their methods not imposing any explicit pruning schedule.      

%% file: sections/method.tex
In this section we detail the pruning methods and heuristics used in this paper. Unless otherwise stated, all experiments involve pruning a ResNet50 network \cite{resnet}, pre-trained on the ImageNet dataset \cite{ILSVRC15} to a top-1 accuracy of $79.02\%$, using the stochastic gradient descent (or SGD) optimizer. The pruning process is conducted over $\mathrm{N}_c = 5$ cycles, each of length $\mathrm{T}_c = 7$ epochs, for a total of $\mathrm{T} = 35$ epochs. 

\subsection{Pruning and Learning-rate Schedules}\label{subsec:schedules}
In this work we either use the $3$-step learning-rate schedule, i.e.,    
\begin{equation}\label{eq:1}
  \mathrm{l}_t =
    \begin{cases}
      1\mathrm{e}-2 & \text{if} \quad 11 > \lfloor t \rfloor \geq 0 \\
      1\mathrm{e}-3 & \text{if} \quad 23 > \lfloor t \rfloor \geq 11\\
      1\mathrm{e}-4 & \text{if} \quad 35 > \lfloor t \rfloor \geq 23
    \end{cases},       
\end{equation}
or the cyclic learning-rate schedule, i.e.,
\begin{equation}\label{eq:2}
  \mathrm{l}_t =
    \begin{cases}
      1\mathrm{e}-2 & \text{if} \quad \lfloor t \rfloor \bmod \mathrm{T}_c 
        \in \{ 0, 1, 2 \} \\
      1\mathrm{e}-3 & \text{if} \quad \lfloor t \rfloor \bmod \mathrm{T}_c 
        \in \{ 3, 4 \} \\
      1\mathrm{e}-4 & \text{if} \quad \lfloor t \rfloor \bmod \mathrm{T}_c 
        \in \{ 5, 6 \} \\
    \end{cases},       
\end{equation}
where $ \mathrm{T} > t \geq 0$ is the batch index normalized by the number of batches per epoch ($\lfloor t \rfloor$, hence, simply denotes the epoch index minus one). The cyclic learning-rate schedule is similar to the learning-rate rewinding scheme of \cite{renda2020comparing}, albeit with a much shorter cycle length.

As for the pruning schedule, we either use the linear, i.e.,
\begin{equation}\label{eq:3}
    \mathrm{r}_t^{\textbf{Lin}} = 1 - (1 - \mathrm{R}_f) \cdot \frac{t}{T},
\end{equation}
or the exponential schedule, i.e.,
\begin{equation}\label{eq:4}
    \mathrm{r}_t^{\textbf{Exp}} = \mathrm{R}_f + \big (1 - \mathrm{R}_f \big ) \cdot \mathrm{exp}(\frac{-t}{\tau}),
\end{equation}
where $\mathrm{r}_t$ and $\mathrm{R}_f$ denote the current and the final target keep-ratios, respectively, and $\tau \ll \mathrm{T}$ is the time-constant for the exponential scheduler. We use a batch size of $128$ and update $\mathrm{r}_t$ every $100$ batches. We note that while the linear schedule prunes the redundant weights at a constant rate, i.e., 
\begin{equation}\label{eq:5}
    \frac{\mathrm{d}}{\mathrm{d}t}(\mathrm{r}_t^{\textbf{Lin}} - \mathrm{R}_f) = -\frac{1 - \mathrm{R}_f}{\mathrm{T}},
\end{equation}
it has an unbounded \emph{normalized} rate of pruning as $t$ approaches $\mathrm{T}$, i.e.,
\begin{equation}\label{eq:6}
    \frac{1}{(\mathrm{r}_t^{\textbf{Lin}} - \mathrm{R}_f)} \cdot \frac{\mathrm{d}}{\mathrm{d}t}(\mathrm{r}_t^{\textbf{Lin}} - \mathrm{R}_f) = -\frac{1}{\mathrm{T} - t}.
\end{equation}
On the other hand, while the exponential schedule imposes a larger rate of pruning at the beginning, its normalized rate of pruning is constant throughout the process, i.e.,
\begin{equation}\label{eq:7}
    \frac{1}{(\mathrm{r}_t^{\textbf{Exp}} - \mathrm{R}_f)} \cdot \frac{\mathrm{d}}{\mathrm{d}t}(\mathrm{r}_t^{\textbf{Exp}} - \mathrm{R}_f) = -\frac{1}{\tau}.
\end{equation}
The intuition that a network is most tolerant of pruning at the beginning when it has a lot of redundant weights (i.e., $\mathrm{r}_t - \mathrm{R}_f$ is large) and least tolerant toward the end when it is the leanest (i.e., $\mathrm{r}_t - \mathrm{R}_f$ is small), motivates the exponential pruning schedule. This intuition also motivates the cyclic learning-rate schedule as it supports the constant normalized rate of pruning by providing a full range of learning-rates throughout the pruning process. To further match the pruning and learning-rate schedules, we use a variant of the exponential schedule whenever used in conjunction with a cyclic learning-rate, where the target keep-ratio is only updated during the first two epochs in each cycle with the objective that most weights are removed during a high learning-rate phase (c.f., the target keep-ratio plot in Figure \ref{fig:lr_rewinding} showing the two variants when used with the $3$-step and cyclic learning-rates).  

\subsection{Unstructured GMP and Random Pruning}
In this work we use the iterative GMP \cite{han2015pruning} and the random pruning methods to demonstrate the implications of cascade weight shedding for unstructured pruning. In both methods we use a threshold, initialized to a small value, e.g., $1\mathrm{e}-4$, to detect and remove degenerate weights. In other words, once the magnitude of a weight drops below the threshold, it is removed and counted toward the \emph{actual} keep-ratio, i.e., $\rho_t$. In both methods we use a pruning schedule to update the target keep ratio, i.e., $\mathrm{r}_t$, at regular intervals, e.g., every $100$ batches, and prune additional weights only if $\mathrm{r}_t$ is less than $\rho_t$. For the GMP these would be the weights with the smallest magnitudes whereas for random pruning they are chosen by chance. Also, where for the GMP we update the threshold to the magnitude of the largest weight pruned so far, we do not update it for the random pruning.   

\subsection{Semi-structured GMP}\label{subsect:semi-structure-gmp}
We use the $4 \times 1$ GMP method \cite{block-sparsity-mao} to illustrate the effects of the cascade weight shedding on semi-structured pruning. The $4 \times 1$ GMP is identical to its unstructured counterpart, except that instead of pruning individual weights based on their magnitudes, blocks of weights corresponding to $4$ adjacent input channels (for the same output channel and spatial coordinates) are pruned based on the blocks' $\mathrm{L}_2$ norms. Note that we do not prune the individual weights within the kept blocks, irrespective of their magnitudes. 

To illustrate the sparsity regularization through selective weight decaying, we also study a variant of the $4 \times 1$ GMP where depending on a block's $\mathrm{L}_0$ norm, had the weights within the block were pruned, a different amount of weight decaying is applied to all of its weights. More specifically we apply a multiple of $1\mathrm{e}-4$ as weight decay where the multiplier corresponding to an $\mathrm{L}_0$ norm of $0, 1, 2, 3$ or $4$ is given by $0, 4, 2, 1$ or $0$, respectively. As can be seen, we apply progressively larger weight decays to blocks with more elements already below the threshold in an attempt to regularize such unstructured sparsity.  

%% file: sections/experiments.tex
In this section we detail the experiments we conducted to investigate the cascade weight shedding phenomenon. We start off, however, with a brief discussion of the effects of pruning and learning-rate schedules to facilitate the presentation.

\subsection{Effects of Pruning and Learning-rate Schedules}\label{subsec:exp_pruning}
\begin{figure}[t]
    \begin{center}
        \includegraphics[width=1.0\linewidth]{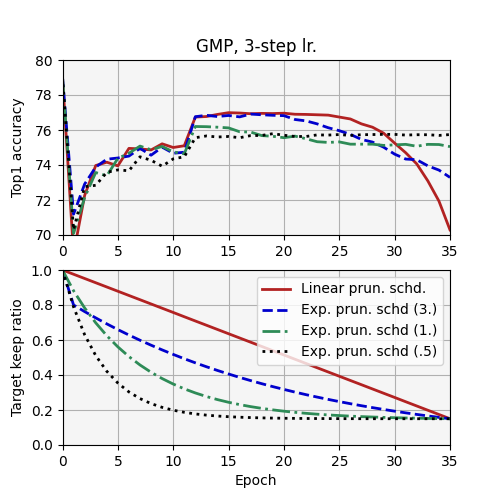}
    \end{center}
   \caption{Effect of pruning schedule, c.f. section \ref{subsec:exp_pruning}.}
\label{fig:exp_pruning}
\end{figure}

Figure \ref{fig:exp_pruning} shows the effects of pruning schedule when compressing a ResNet50 trained on ImageNet using GMP. As can be seen, both the linear schedule and the exponential schedule with the time-constant of $3$ perform better than the more aggressive ones initially, i.e., $1\%$ between epochs $15$ to $20$, only to lose out toward the end. This validates our intuition in section \ref{subsec:schedules} that the last percentage points shed off are the most challenging ones and that the exponential schedule, for which the normalized rate of change is constant, c.f., \eqref{eq:7}, is more suitable. The optimal time-constant for the exponential pruning schedule depends on both the total number of pruning epochs $\mathrm{T}$ and the final keep-ratio $\mathrm{R}_f$. 

\begin{figure}[t]
    \begin{center}
        \includegraphics[width=1.0\linewidth]{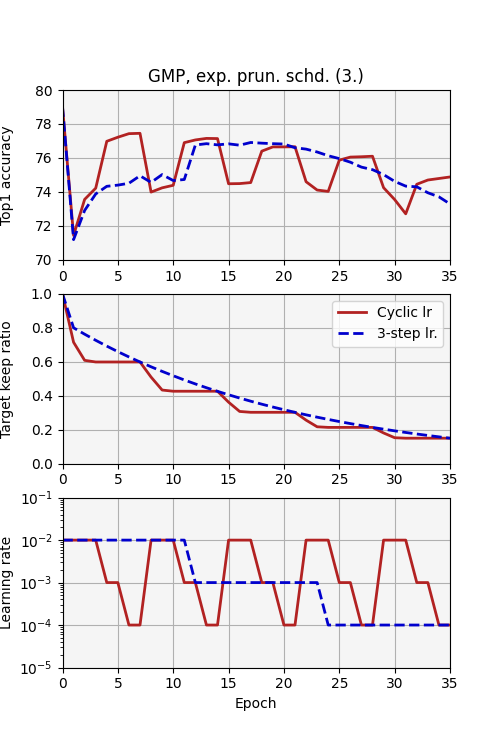}
    \end{center}
   \caption{Effect of learning-rate schedule, c.f., section \ref{subsec:exp_pruning}.}
\label{fig:lr_rewinding}
\end{figure}

A closer examination of Figure \ref{fig:exp_pruning} reveals the great impact of learning-rate on performance. For example the top-1 accuracy jumps by more than one percent on epoch $12$ when the learning-rate is reduced from $1\mathrm{e}{-2}$ to $1\mathrm{e}{-3}$. Even more importantly, all pruning schedules (except for the most aggressive one), start declining at epoch $24$ when learning-rate is further reduced to $1\mathrm{e}{-4}$. In other words the performance hurts because the learning-rate is too small to support the needed adaptation of weights. The cyclic learning-rate schedule alleviates this problem by switching between high, moderate and low learning-rates throughout the pruning process, c.f., Figure \ref{fig:lr_rewinding}.

\subsection{Cascade Weight Shedding}\label{subsec:gmp_shedding}
\begin{figure}[t]
    \begin{center}
        \includegraphics[width=1.0\linewidth]{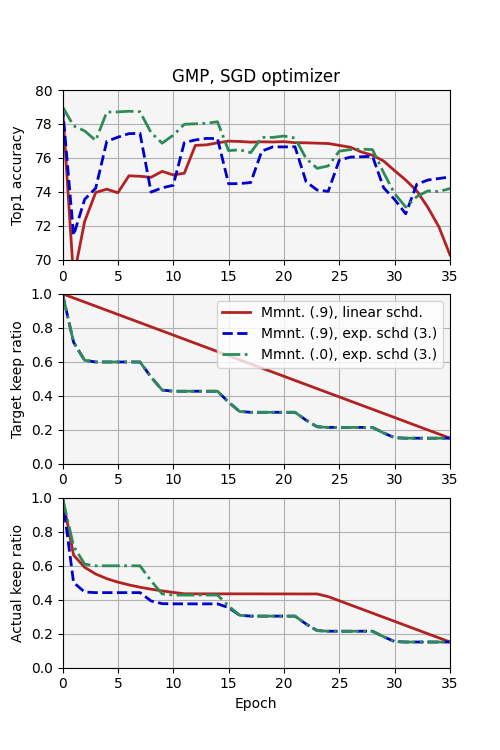}
    \end{center}
   \caption{Effect of cascade weight shedding on GMP, c.f., sections \ref{subsec:gmp_shedding},  \ref{subsec:rand_shedding} and \ref{subsec:weight_rewinding}.}
\label{fig:gmp_shedding}
\end{figure}

Revisiting Figure \ref{fig:exp_pruning} reveals another important aspect, i.e., that the top-1 accuracy suffers a near-catastrophic hit, just as the pruning starts, before quickly recovering a few epochs later (a similar observation was reported in \cite{zhu2017prune}, but went unexplained). To illustrate this further, we compare the GMP pruning results of ResNet50 for three different scenarios in Figure \ref{fig:gmp_shedding}. Note that the two scenarios with exponential pruning schedules are identical except for the SGD optimizer's momentum parameter (e.g., $0$ vs. $0.9$), yet this single difference makes a profound effect, namely, pruning with a momentum of $0$ prevents the initial performance hit. It also allows the actual keep-ratio to follow the target closely, whereas with a momentum of $0.9$ it is at times more that $15\%$ under. The disparity between the target and the actual keep-ratios is even wider under the linear pruning schedule (i.e., $40\%$ at epoch $5$) and is a manifestation of the cascade weight shedding phenomenon.

A closer examination of the actual keep-ratio's time trace shows that the weight shedding starts very early in the pruning process. In fact, it is triggered by the removal of the very first few percentage of weights. It also has a cascading nature where the pruning of the first (seed) weights causes a large number of other weights to be shed, which in turn causes shedding of still other weights, and the process continues. In fact, it is this cascading effect that explains the exponential characteristics of the actual keep-ratio's time-trace (i.e., the bottom plot of Figure \ref{fig:gmp_shedding}), even when a linear (and not exponential) pruning schedule is used.

As to the question of necessary conditions, our experiments show that the cascade weight shedding may occur when pruning a ResNet50 network using an SGD optimizer with a momentum of $0.9$. It may also occur, although to a lesser degree, with a smaller momentum of $0.6$ or $0.3$. Our experiments further show that while cascade weight shedding may occur with an EfficientNet-b0 model, it is much less severe and occurs only at a very high momentum of $0.99$. We have not observed cascade weight shedding in MobileNetV3 even for a momentum of $0.99$. These observations, together with the report from \cite{zhu2017prune} concerning InceptionV3 \cite{inceptionv3} (if it is indeed due to the cascade weight shedding), lead us to speculate that a large (i.e., one with 3D convolutional as opposed to depth/point-wise layers), fully-trained architecture and a high momentum value are essential conditions for the cascade weight shedding to occur.

\subsection{Effects of Cascade Weight Shedding on Unstructured Pruning}\label{subsec:rand_shedding}
The fact that pruning a small number of weights can trigger a cascade process implies that GMP does not operate based on identifying redundant weights, a pillar of unstructured pruning since the start \cite{obd, obs}. It further implies that, in a fully-trained model, pruning a set of weights renders another set of weights unfavorable as they now induce noise. Such weights are subsequently shed via back-propagation. The strength of the (iterative) GMP is that it next prunes these newly shed weights instead of the ones in active use. In this sense, iterative GMP is akin to picking fruits by shaking a tree and collecting the fallen ones. In the case of a large fully-trained network with a high momentum, such weight shedding can trigger a cascade process where a very large number of loose weights are removed very rapidly, similar to an avalanche. The looseness assertion is verified by Figure \ref{fig:gmp_shedding}, which indicates that the top-1 accuracy of the GMP with a momentum of $0.9$ is slightly better than GMP with a momentum of $0$. This, however, does not mean that the same weights are pruned, c.f., section \ref{subsec:which_weights}.

\begin{figure}[t]
    \begin{center}
        \includegraphics[width=1.0\linewidth]{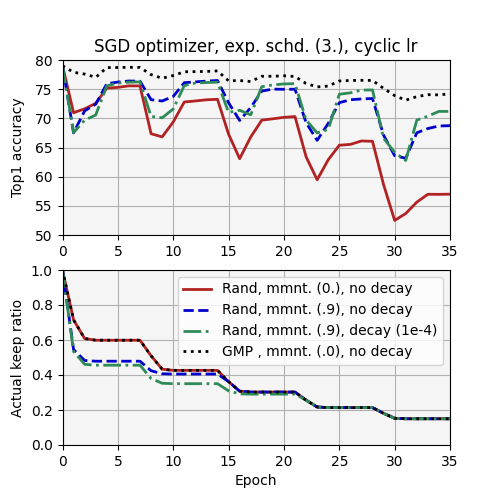}
    \end{center}
   \caption{Effect of weight shedding on Random pruning, c.f., section \ref{subsec:rand_shedding}.}
\label{fig:rand_shedding}
\end{figure}

As further evidence for the hypothesis that cascade weight shedding only removes loose weights, Figure \ref{fig:rand_shedding} compares the performance of the random pruning method in the presence and absence of cascade weight shedding (i.e., momentum values of $0.9$ and $0$, respectively). As the figure indicates, random pruning in the absence of cascade weight shedding performs quite poorly which is not surprising. What is surprising, however, is the $11.7\%$ boost that it gets from cascade weight shedding. This is because in the latter case most of the weight removals are due to the cascade shedding rather than the random pruning. The performance can further be improved by applying a small amount of weight decay (i.e., $1\mathrm{e}{-4}$), which helps pushing down the already shed weights further below the threshold (GMP prunes these weights by design, without any need for weight decaying). Figure \ref{fig:rand_shedding} also illustrates a potential pitfall of cascade weight shedding, namely, concealing a pruning method's inherent weaknesses. For example it gives the impression that random pruning, a very poor method, lags GMP only by a few percentage points, i.e., $3.7\%$, and even outperforms it by $0.9\%$, had a linear pruning schedule been used for the latter.

Revisiting Figure \ref{fig:gmp_shedding} reveals a second potential pitfall of cascade weight shedding when pruning networks. As the actual keep-ratio plot shows, both the rate of weight shedding and its extent vary and depend on various aspects of the triggering event such as the seed weights, their count and how quickly they are pruned. This means that if the triggering event is too severe or the target keep-ratio is too high (e.g., $0.6$), the actual keep-ratio may stabilize at a level below the final target keep-ratio, leaving behind a large number of degenerate weights. This makes the accuracy suffer compared to that in the absence of cascade weight shedding notwithstanding which weights have been shed. Also, the very high rate of shedding, e.g., $55\%$ of the entire network in less than $5$ epochs for the linear schedule, makes designing pruning methods challenging as it implies that the relevance of weights changes rapidly and significantly throughout the pruning process and that network statistics are highly non-stationary. This high rate of weight shedding also presents an opportunity for speeding-up the GMP method without sacrificing its performance, through a combination of a proper triggering event, learning-rate and pruning schedules.

\subsection{Possible connections to Weight and Learning-rate Rewinding}\label{subsec:weight_rewinding}
The cascade weight shedding may in fact explain the advantage of weight and learning-rate rewinding schemes over fine-tuning. For example, \cite{renda2020comparing} reports that fine-tuning a ResNet50 with a momentum of $0.9$, after pruning $20\%$ of its weights achieves a lower accuracy compared to retraining by weight rewinding, where the kept weights and the learning-rate are first re-wound to their values from earlier in training. It further reports that rewinding only the learning-rate has a similar, if not better effect. As Figure \ref{fig:gmp_shedding} suggests, pruning in this scenario most likely results in the shedding of a significant percentage of the weights beyond the initial, e.g., $20\%$ seed in a few short epochs. This high rate of shedding requires larger learning-rates than those normally used with fine-tuning to support the required weight updates.

Interestingly, rewinding weights makes the model "less fully-trained", hence less prone to cascade weight shedding, which alleviates the need for using higher learning-rates. If cascade weight shedding is indeed the root cause for the under performance of fine-tuning, other remedies beside weight and learning-rate rewinding may be possible. These include fine-tuning with a momentum of $0$ to prevent the cascade weight shedding altogether, or even keeping the original momentum of $0.9$ but pruning the, e.g., $20\%$ seed weights gradually over a few epochs to reduce weight shedding's extent. It also follows that weight or learning-rate rewinding would not show advantage over fine-tuning for networks not prone to cascade weight shedding such as EfficientNet-b0 or MobileNetV3. Obviously, the speculated connection needs a thorough validation which is beyond the scope of this work.

\subsection{Which Weights are Pruned?}\label{subsec:which_weights}
\begin{figure}[t]
    \begin{center}
        \includegraphics[width=1.0\linewidth]{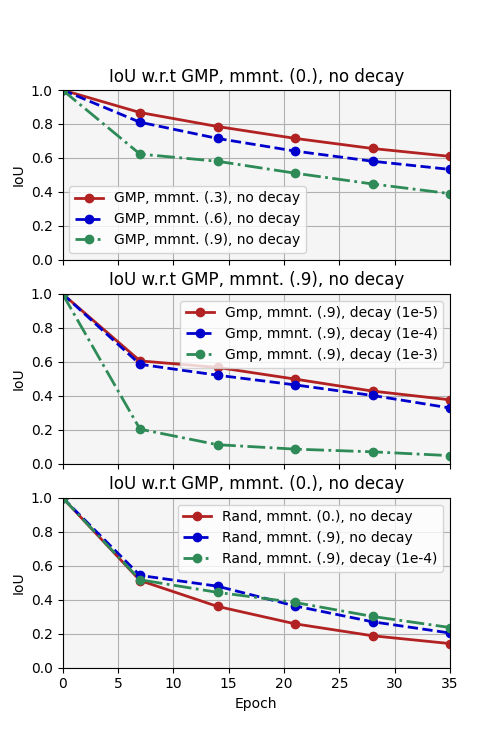}
    \end{center}
   \caption{Effects of cascade weight shedding and weight decaying on the set of kept weights, c.f., section \ref{subsec:which_weights}.}
\label{fig:iou_1x1}
\end{figure}

Figure \ref{fig:iou_1x1} provides insight into the role of cascade weight shedding in determining which weights are kept, i.e., it gives the intersection over union, i.e., IoU, for the set of kept weights under two scenarios. The top plot shows that the GMP method with momentum values of $0$ and $0.9$ (with similar performances) have an IoU of $0.4$ at the final keep-ratio of $0.15$. This means that different weights are pruned depending on the presence of cascade weight shedding and its extent. The middle plot indicates a similar effect with respect to weight decaying, i.e., that moderate values of weight decay, with similar performances, have an IoU around $0.4$. An aggressive weight decay of $1\mathrm{e}-3$, with a much lower top-1 accuracy, results in a considerably smaller IoU, i.e., $0.05$. The bottom plot shows a similar trend too where random pruning with a momentum of $0$, i.e., a poor pruning method, yields a small IoU of $0.15$. Interestingly, improving random pruning's performance through cascade weight shedding (by using a momentum of $0.9$) pushes the IoU above $0.2$. This suggests that methods with a good performance tend to keep a common set of weights, though the overall set of kept weights varies from one method to the next, and is affected by, e.g., cascade weight shedding and weight decaying.

\subsection{Effects of Cascade Weight Shedding on Semi-structured Pruning}\label{subsec:semi_struct}
\begin{figure}[t]
    \begin{center}
        \includegraphics[width=1.0\linewidth]{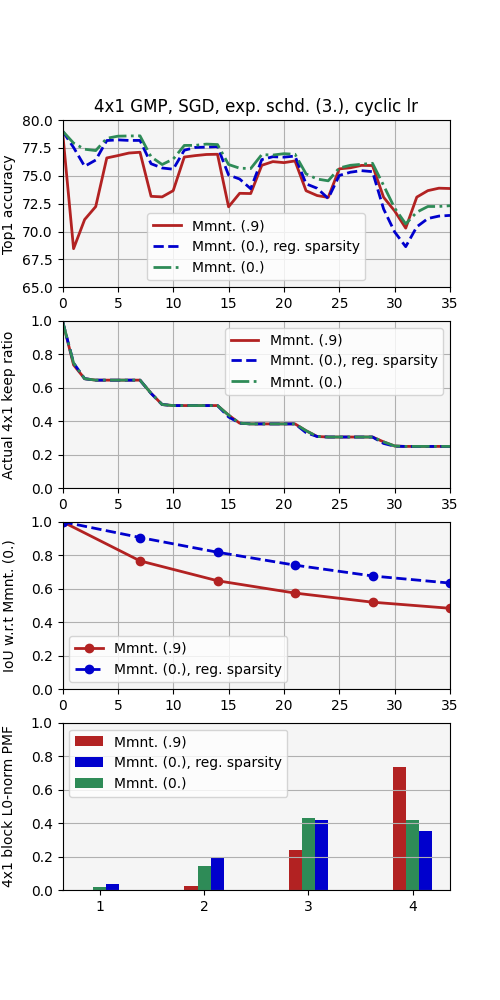}
    \end{center}
   \caption{Effects of cascade weight shedding and weight decaying on 4x1 semi-structured GMP, c.f., section \ref{subsec:semi_struct}.}
\label{fig:semi_struct}
\end{figure}

The fact that cascade weight shedding can influence which weights are kept is of great importance for semi-structured pruning, where it is desired to regularize the induced sparsity according to certain patterns \cite{block-sparsity-mao, block-sparsity-narang, balanced-sparsity-yao, balanced-sparsity-cao}. Figure \ref{fig:semi_struct} compares the $4 \times 1$ semi-structured GMP \cite{balanced-sparsity-yao} scenarios with (i.e., momentum of $0.9$) and without (i.e., momentum of $0$) cascade weight shedding. The occurrence of the cascade weight shedding for the former scenario is confirmed by the initial drop in its top-1 accuracy. The fact that the target and the actual keep-ratios, despite this occurrence, fall on top of one another is not surprising as these are \emph{block} keep-ratios. The IoU plot also shows that, as expected, the sets of kept blocks differ significantly between the two scenarios. The main observation here is that cascade weight shedding improves the final top-1 accuracy by $1.5\%$ (in practice the pruned models will be further fine-tuned, so the difference may be a bit different). This is significant, especially noting that regularization through selective weight decaying (as detailed in section \ref{subsect:semi-structure-gmp}) actually hurts the performance by $0.9\%$. The fact that weight decaying degrades GMP's performance is not surprising, as it skews the classification loss without benefiting it as argued in \ref{subsec:rand_shedding}. To illustrate why cascade weight shedding improves semi-structured GMP, the bottom plot gives the PMF of the kept blocks' $\mathbf{L}_0$ norm \emph{assuming} that weights with a magnitude less than $6\mathrm{e}-3$ are degenerate (repeating with other values gives qualitatively similar results). As the plot shows, cascade weight shedding allows the kept blocks to have a significantly lower percentage of degenerate weights, i.e., $25\%$ compared to $60\%$. This again is expected as cascade weight shedding nulls out a large number of degenerate weights early in the process, enabling the semi-structured GMP to prune out the blocks with the most degenerate ones and keep the ones that have the least number of them. We expect the cascade weight shedding to have a similar enhancing effect with respect to balanced-sparsity variants \cite{balanced-sparsity-yao}.    

%% file: LaTeX/sections/future_directions.tex
As was shown, cascade weight shedding can benefit pruning by improving its performance and reducing its computational complexity, or hurt it by concealing its inherent weaknesses and leaving behind too many degenerate weights. A better understanding of the various aspects of the cascade weight shedding is hence critical. Such aspects include conditions necessary for its occurrence such as architecture characteristics, type of optimizer, amount of training, learning-rate and pruning schedules, training dataset, characteristics of the triggering event that control the final extent of shedding, and factors that enable efficient regularization of the induced sparsity as desired.   

Another potential application of the weight shedding phenomenon (whether controlled, or cascading) is with respect to identifying the sets of co-dependent weights, sometimes referred to as the winning lottery tickets \cite{lotteryticket, deconstructinglottery}. This may be done in two ways. One is by knocking out a small set of seed weights and studying the set of weights subsequently shed. Another possibility is through repeated pruning of a network using, e.g., different sets of seed weights, weight decays or momentum values, and analyzing the common subset of kept weights among them. 

%% file: sections/conclusion.tex
We reported, for the first time, on the cascade weight shedding phenomenon in deep neural networks, where in response to pruning a small percentage of a network's weights, a large percentage of the remaining is rapidly shed during the ensuing fine-tuning phase. We showed that cascade weight shedding, when present, can significantly improve the performance of an otherwise sub-optimal scheme such as random pruning. This explains why some pruning methods may perform well under certain circumstances, but poorly under others. We provided insight into why iterative GMP despite its simplicity, provides such competitive performances for a wide range of scenarios. We also demonstrated cascade weight shedding's potential for improving GMP's accuracy, and speeding it up, hence alleviating its computational complexity. In doing so, we highlighted the importance of pruning and learning-rate schedules. We shed light on weight and learning-rate rewinding methods of re-training. In particular, we discussed their possible connections to cascade weight shedding, explaining the reason for their advantage over fine-tuning. We investigated cascade weight shedding's effects on which weights are pruned, and its implications for semi-structured pruning. Finally we provided a few directions for future research.